\newif\ifanonymous
\anonymousfalse

\documentclass[letterpaper, 10 pt, conference]{ieeeconf}

\ifanonymous\else
\IEEEoverridecommandlockouts
\fi

\usepackage{graphicx} 
\usepackage{booktabs} 
\usepackage{multirow} 
\usepackage{lipsum} 
\usepackage{lettrine} 
\usepackage{xcolor} 
\usepackage{cite} 
\usepackage{amsmath} 
\usepackage{amssymb}  
\usepackage{hyperref} 
\usepackage{cleveref} 
\usepackage{xfrac} 
\usepackage[caption=false,font=footnotesize]{subfig}
\usepackage{algorithm}
\usepackage{algorithmic}
\usepackage{bm} 
\usepackage{siunitx} 

\newcommand{\citetodo}[1]{\textcolor{red}{[?]}}

\newcommand{\opt}{\ensuremath{^\star}} 
\newcommand{\refa}{\ensuremath{^{\dagger}}} 
\newcommand{\Tr}{{\scriptsize\textsc{(Tr)}}} 
\newcommand{\Val}{{\scriptsize\textsc{(Val)}}}  
\newcommand{\Vl}{{\scriptsize\textsc{(Vl)}}}  

\definecolor{scarRed}{RGB}{200, 55, 113}
\definecolor{scarGreen}{RGB}{70, 124, 122}
\definecolor{scarViolet}{RGB}{126, 87, 194}

\title{\LARGE \bf
SCAR: Satellite Imagery-Based Calibration for Aerial Recordings
}

\ifanonymous
    \author{Anonymous Authors}
\else
    \author{Henry Hölzemann$^{1}$ and Michael Schleiss$^{2}$
    \thanks{$^{1}$Henry Hölzemann is with the Department of Sensor Data and Information Fusion,
        Fraunhofer FKIE, 53343 Wachtberg, Germany
        {\tt\small henry.hoelzemann@fkie.fraunhofer.de}}%
    \thanks{$^{2}$Michael Schleiss is with the Chair of Astronautics, University of the Bundeswehr Munich,
        85579 Neubiberg, Germany
        {\tt\small michael.schleiss@unibw.de}}%
    \thanks{Code will be available at {\tt\small github.com/hlzmnhnry/SCAR}}
    }
\fi

\begin{document}

\maketitle

\thispagestyle{empty}
\pagestyle{empty}

\begin{abstract}


We introduce SCAR, a method for long-term auto-calibration refinement of aerial visual-inertial systems that exploits georeferenced satellite imagery as a persistent global reference. 
SCAR estimates both intrinsic and extrinsic parameters by aligning aerial images with 2D--3D correspondences derived from publicly available orthophotos and elevation models. 
In contrast to existing approaches that rely on dedicated calibration maneuvers or manually surveyed ground control points, 
our method leverages external geospatial data to detect and correct calibration degradation under field deployment conditions. 
We evaluate our approach on six large-scale aerial campaigns conducted over two years under diverse seasonal and environmental conditions.
Across all sequences, SCAR consistently outperforms established baselines (Kalibr, COLMAP, VINS-Mono), reducing median reprojection error by a large margin, 
and translating these calibration gains into substantially lower visual localization rotation errors and higher pose accuracy. 
These results demonstrate that SCAR provides accurate, robust, and reproducible calibration over long-term aerial operations without the need for manual intervention.

\end{abstract}

\section{INTRODUCTION}

\lettrine{A}{ccurate} sensor calibration is fundamental for enabling reliable visual-inertial odometry (VIO) and visual localization (VL).
These technologies are key enablers for autonomous systems operating in environments with limited access to the Global Navigation Satellite System (GNSS), such as urban canyons, forests, and interference-prone regions, where vision-based localization and navigation allow unmanned aerial vehicles (UAVs) to maintain robust operation.
This capability is critical for tasks such as search and rescue, autonomous aerial delivery, or disaster response, where autonomous flight must remain robust over extended periods of time. \par

To obtain robust ego-motion estimates from visual odometry (VO), an accurate calibration of camera-specific intrinsic parameters is required.
When these estimates are fused with measurements from an inertial navigation system (INS), an additional extrinsic calibration between the camera and the INS becomes crucial.
Over the past years, robust toolboxes and methods have been developed and established for estimating both intrinsic camera parameters and the extrinsic calibration between the camera and INS \cite{heng2013camodocal, furgale2013unified}.
Nevertheless, most existing approaches rely on manual calibration with dedicated targets or on specialized datasets that often diverge from practical deployment scenarios, making them likely to cause inaccurate georeferencing results in application scenarios.
In many real-world aerial applications, GNSS is not permanently unavailable but intermittently accessible, for example during high-altitude flight or in open areas.
These GNSS-enabled phases are typically used to initialize and maintain navigation systems, while reliable visual–inertial operation is required during subsequent GNSS outages.
Maintaining accurate calibration during GNSS availability therefore directly improves robustness in GNSS-denied scenarios, and remains beneficial even in fully GNSS-enabled operation through sensor fusion.
This becomes particularly critical in autonomous deployments over extended periods, such as aerial delivery services or inspection fleets, where sensors undergo gradual changes due to mechanical stress, re-mounting, or environmental conditions.

Small miscalibrations may be tolerable at low altitudes, but they accumulate to significant errors in high-altitude flights, severely degrading localization accuracy.
Another relevant scenario is large-scale manufacturing: while factory calibration provides initial parameters, each individual system typically requires additional manual refinement before deployment.
Without automatic recalibration, such systems would therefore require repeated manual interventions, which might be impractical at scale. \par

For these reasons, we present SCAR as an approach to automatic long-term calibration refinement. 
SCAR does not operate as purely internal self-calibration, but instead validates and adapts intrinsics and extrinsics against external geospatial references, directly optimizing the reprojection consistency in a georeferenced environment. 
Leveraging publicly available satellite data, it provides stable anchors across seasons and flight campaigns, independent of motion patterns or calibration infrastructure, and scales naturally to repeated deployments or even entire UAV fleets. 
As a result, calibration can be maintained across long time periods without manual intervention, and the framework additionally enables retroactive recalibration from previously collected flight data without the need for dedicated calibration sequences. 
By addressing these challenges, the main contributions of this paper are threefold:

\begin{enumerate}
    \item \textbf{Novel problem perspective:} We view long-term calibration as repeated validation and refinement against automatic geospatial references, introducing a formulation that jointly verifies and optimizes parameters in the georeferenced domain.
    \item \textbf{Open-source framework:} We provide SCAR as a modular toolbox that makes our approach accessible to the community, enabling reproducible evaluation and refinement of intrinsics and extrinsics. \newpage
    \item \textbf{Multi-year evaluation on real-world data}: We demonstrate our method on aerial datasets collected over multiple years, showing that it mitigates calibration drift and improves robustness of long-term deployments compared to state-of-the-art methods.
\end{enumerate}

\begin{figure*}[t]
  \centering
  \includegraphics[width=\textwidth]{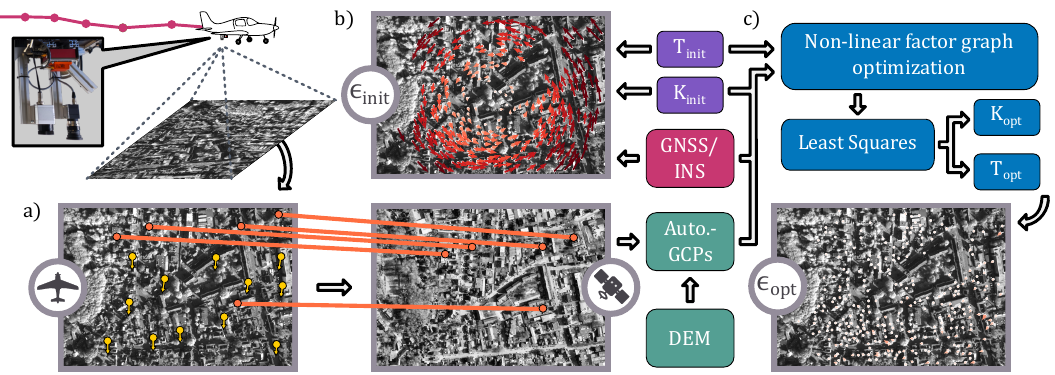}
  \caption{Overview of the proposed SCAR framework for long-term INS–camera calibration refinement. (a) Aerial imagery is matched against satellite or orthophoto data to generate georeferenced anchors, while feature correspondences are tracked across frames to provide multi-view constraints for optimization. (b) Initial intrinsic and extrinsic calibrations are used to evaluate reprojection errors of these anchors. (c) Both absolute anchors and absolute pose measurements (GNSS/INS) are fused in a non-linear factor graph optimization, yielding refined intrinsic and extrinsic calibration parameters.}
  \label{fig:introduction_overview}
\end{figure*}

The remainder of this paper is organized as follows: \Cref{section:related-work} reviews related work. \Cref{section:scar-framework} introduces our framework. \Cref{section:experiments-and-discussion} describes the experimental setup and presents the results. Finally, \cref{section:conclusion-and-future-work} concludes the paper.


\section{RELATED WORK}
\label{section:related-work}

The common practice in visual–inertial sensor calibration relies on offline procedures in controlled environments.
These methods are often target-based \cite{furgale2013unified, nikolic2016non, zhang2002flexible, mirzaei2008kalman} and are typically conducted at short ranges and limited scales, which makes them less representative of actual operating conditions in high-altitude scenarios.
Alternatively, so-called auto-calibration methods aim to eliminate manual intervention by estimating parameters directly from natural scene observations and sensor motion, typically within Structure-from-Motion (SfM) pipelines. 
While avoiding dedicated targets, such methods remain computationally demanding and are generally restricted to offline use \cite{schonberger2016structure, baid2023distributed}.

To overcome these limitations, numerous online methods have been developed \cite{kelly2011visual, jones2011visual, yang2016monocular, huang2019visual}.
Existing approaches mostly build on relative motion and are often integrated into VIO or SLAM frameworks \cite{leutenegger2015keyframe, yang2016self, schneider2017visual, huang2018online, qin2018vins}, where extrinsic parameters are jointly estimated as part of the optimization --- commonly through bundle adjustment.
Such online calibration methods can adapt to time-varying misalignments and correct residual errors left by offline procedures, thereby improving long-term consistency during operation.
In addition to extrinsic estimation, a smaller body of work has also explored online refinement of camera intrinsics \cite{huang2021camera, kersting2011mounting, gneeniss2015flight}, demonstrating the feasibility of self-calibration during operation. \par

In the specific context of aerial robotics, additional sources of absolute reference have been considered for calibration.
Ground control points provide precise geodetic anchors by linking image observations to surveyed landmarks \cite{mostafa2001boresight, honkavaara2004flight}, but their preparation is costly and spatially limited.
GNSS and INS measurements are standard on many UAV payloads, offering absolute constraints that can help mitigate drift over calibration sequences \cite{kersting2011mounting, huang2021camera, guo2023enhancing}.
However, even with GNSS/INS integration, subtle camera–INS misalignments often remain unresolved, as these measurements primarily constrain position and orientation at a coarser level.
This highlights the need for complementary georeferenced information that directly links image observations to the environment.\par

Beyond GNSS and INS measurements, and traditional GCPs, recent work leverages visual localization against aerial or satellite imagery to provide absolute reference information \cite{couturier2021review, patel2020visual, qiu2025high}.
By performing image matching between UAV images and georeferenced maps, such methods effectively create “automatic GCPs” at scale, avoiding costly ground surveys.
These matches provide globally consistent anchors that naturally complement GNSS for calibration purposes, as the camera information is directly incorporated, capturing misalignments that GNSS/INS measurements alone cannot reveal.
Despite this potential, aerial-to-satellite image matching has rarely been exploited for visual–inertial sensor calibration, leaving an open opportunity for long-term, drift-free refinement. \par

Most closely related to our approach, Bender et al.~\cite{bender2014ins} perform INS–camera calibration without relying on GCPs, formulating the problem as a factor-graph optimization. 
More recently, Guo et al.~\cite{guo2023enhancing} enhance boresight calibration by incorporating coplanarity and relative height constraints from digital elevation models (DEMs). 
While such methods reduce reliance on surveyed landmarks, they remain limited to local consistency within a single deployment. 
Existing online calibration approaches, although capable of adapting parameters during operation, also depend primarily on relative motion and therefore lack absolute reference, which restricts reproducibility across different environments and long-term campaigns. 
In contrast, SCAR explicitly integrates GNSS/INS priors with automatically derived image-based anchors in the world frame, providing two independent and complementary sources of global reference. 
This combination enables stable refinement of intrinsics and extrinsics and ensures consistent calibration across deployments, beyond the scope of current methods.

\section{THE SCAR FRAMEWORK}
\label{section:scar-framework}

\subsection{Method Overview}
\label{subsection:method-overview}

In this section, we present the SCAR framework, designed to achieve long-term refinement of both INS–camera extrinsics and camera intrinsics.
SCAR operates on existing intrinsic and extrinsic calibration parameters, which may be obtained from prior offline calibration or coarse factory values, and refines them during long-term operation.
Our approach uniquely integrates georeferenced 6-DoF poses and image-based anchors into a unified optimization pipeline.
We begin by providing a high-level overview of the inputs, core steps, and outputs of our method, before describing the individual components in detail.
The overall workflow of the proposed SCAR framework is summarized in \Cref{fig:introduction_overview}. \par

Given the absolute position and orientation of a UAV at each image capture time from GNSS/INS, a spatially aligned satellite or orthophoto cutout is extracted.
Images taken on board the UAV by a downward-facing camera are matched against this reference to obtain 2D–2D correspondences.
By incorporating DEM information, these satellite correspondences are lifted to approximate 3D georeferenced anchors, which in turn provide reprojection constraints in the UAV imagery.
Image features are tracked across overlapping frames to ensure that these anchors are observed repeatedly over time within the aerial view, creating multi-frame observations that provide constraints required for bundle adjustment and improve robustness within the optimization. \par

All measurements --- GNSS/INS pose priors, georeferenced anchors, pixel projections and initial intrinsic and extrinsic calibration parameters --- are combined in a nonlinear factor-graph optimization.
Each measurement is incorporated with its respective uncertainty, ensuring that the different absolute information sources are weighted appropriately during optimization.
To improve convergence stability, we employ a staged optimization schedule in which different parameter groups are optimized sequentially while all others are held fixed, concluding with a final joint refinement of all variables.
This staged procedure reduces degeneracies and prevents the optimizer from converging to poor local minima.
For this work, we assume a near-nadir camera configuration, which simplifies anchor generation by reducing geometric ambiguities.
While SCAR is presented in this setting, the approach can be adapted to oblique camera configurations with appropriate modeling.
Finally, a dedicated least-squares adjustment is performed to transfer systematic discrepancies from the optimized camera trajectory into updated extrinsic calibration parameters.
This separation of graph-based refinement and extrinsic adjustment ensures both robustness during optimization and interpretability of the final calibration result.
We next describe the modeling of GNSS/INS pose priors, which form the backbone of our optimization.

\subsection{Pose Prior Modeling}
\label{subsection:pose-prior-modeling}

For a camera image $I_t$ at time $t$, we assume an associated pose measurement $\hat P^W_{\text{INS},t} = (\hat R^W_{\text{INS},t}, \hat p^W_{\text{INS},t}) \in SE(3)$ in the world coordinate system $W$, provided by an onboard INS.
This navigation solution fuses GNSS with inertial and potentially additional aiding sensors, yielding a position $\hat p^W_{\text{INS},t}$ in the Universal Transverse Mercator (UTM) projection and an orientation $\hat R^W_{\text{INS},t}$ expressed in East–North–Up (ENU).

Given an initial extrinsic calibration $T^{\text{INS}\to\text{CAM}}_{\text{init}} \in SE(3)$, we obtain an approximate camera pose for each image as
$$
\hat P^W_{\text{CAM},t} = \hat P^W_{\text{INS},t}\, T^{\text{INS}\to\text{CAM}}_{\text{init}}.
$$
This extrinsic transform is expected to contain errors and is itself subject to refinement in later stages of the SCAR framework.
Yet, for robustness, we directly optimize camera poses in the factor graph rather than treating INS poses and extrinsics as separate variables; the refined extrinsic parameters are subsequently recovered from optimized poses via a dedicated least-squares adjustment (see \Cref{subsection:calibration-refinement}).

Since optimized camera poses are expected to remain close to the initial estimates, we formulate pose priors as residuals:
$$
r_t^{\text{rot}} = \operatorname{Log}\!\left(\hat R^W_{\text{CAM},t} \, (R^W_{\text{CAM},t})^\top\right),
\quad
r_t^{\text{pos}} = \hat p^W_{\text{CAM},t} - p^W_{\text{CAM},t}, 
$$
where $C_t = (R^W_{\text{CAM},t},p^W_{\text{CAM},t})$ is the estimated camera pose and $\operatorname{Log}: SO(3)\to\mathbb{R}^3$ is the Lie algebra log map. Together, these terms define a 6D residual $r_t^{\text{cam}} = [r_t^{\text{rot}}; r_t^{\text{pos}}] \in \mathbb{R}^6$.

Each residual is weighted according to its uncertainty, which combines contributions from three sources: (i) measurement noise of the GNSS/INS solution $\Sigma_t^{\text{GNSS/INS}}$, (ii) uncertainty in the provided extrinsic calibration $\Sigma^{\text{calib}}$, and (iii) a lumped term $\Sigma_t^{\text{lump}}$ capturing unmodeled effects such as temporal misalignment and small temporal extrinsic errors. The resulting covariance for the pose prior is modeled as
$$
\Sigma_t^{\text{cam}} = \Sigma_t^{\text{GNSS/INS}} + \Sigma^{\text{calib}} + \Sigma_t^{\text{lump}}
\;\approx\;
\mathrm{diag}(\sigma_{C_t,r}^2 I_3,\, \sigma_{C_t,p}^2 I_3),
$$
with $\sigma_{C_t,r}$ and $\sigma_{C_t,p}$ denoting characteristic rotational and positional standard deviations, respectively. The contribution of each pose prior to the overall objective is then given by
$$
\| r_t^{\text{rot}} \|^2_{\Sigma_t^{\text{rot}}} \;+\; \| r_t^{\text{pos}} \|^2_{\Sigma_t^{\text{pos}}}.
$$
While GNSS/INS-based pose priors provide absolute constraints for each image, they alone are insufficient to resolve subtle calibration errors.
We therefore complement them with image-based anchors derived from aerial-to-satellite matching, as described next.

\subsection{Aerial-to-Satellite Matching and Anchor Generation}

\begin{figure}[b]
  \centering
  \includegraphics[width=0.95\columnwidth]{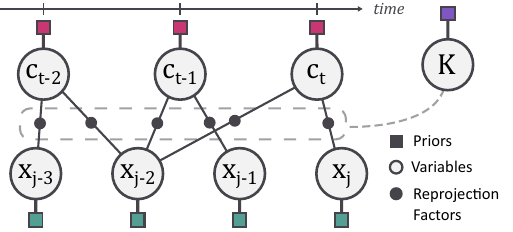}
  \caption{Factor graph structure of SCAR. Camera poses $C_t$, anchors $X_j$, and intrinsics $K$ are modeled as variables, constrained by  \textcolor{scarRed}{pose priors} $\hat P^W_{\text{CAM},t}$, \textcolor{scarGreen}{anchor priors} $\hat{X}^W_j$, \textcolor{scarViolet}{calibration prior} $K_\text{init}$ and reprojection factors.}
  \label{fig:scar_framework_factor_graph}
\end{figure}

We generate geodetic anchors by aligning UAV imagery with georeferenced satellite cutouts.
By combining these correspondences with DEM data, 3D world points are obtained.

For each aerial image $I_t$, a spatially aligned satellite cutout is extracted based on the camera footprint estimated from the INS pose and DEM altitude. 
An image matching procedure (see \Cref{section:experiments-and-discussion}) produces 2D–2D correspondences
$$
(u_{t,j}, v_{t,j}) \in \mathbb{R}^2 \times \mathbb{R}^2,
$$
where $u_{t,j}$ denotes a pixel in the UAV image and $v_{t,j}$ a corresponding pixel in the satellite cutout, both associated with landmark $j$.
Outliers are removed by robust homography estimation.
Each satellite pixel $v_{t,j} = (x,y)$ is associated with an elevation $h(x,y)$ from the DEM, producing a 3D georeferenced world point
$$
\hat{X}^W_j = \big(\pi^{-1}(v_{t,j}), h(x,y)\big) \in \mathbb{R}^3,
$$
where $\pi^{-1}$ denotes the inverse projection from image to world coordinates based on corresponding satellite data.
For each world point, we formulate the residual
$$
r_j^{\text{anc}} = \hat{X}^W_j - X_j,
$$
where $X_j$ is the optimization variable for landmark $j$.
The resulting anchor couples this world point with its UAV observation $a_{t,j} = (X_j, u_{t,j})$.
To obtain multi-frame observations, the UAV pixel $u_{t,j}$ from $t=t_0$ is tracked across subsequent frames $\{t_1, \dots, t_k\}$, yielding a set of anchors for $X_j$:
$$
a_j = \{(X_j, u_{t_0,j}), \dots, (X_j, u_{t_k,j})\}.
$$
These observed anchors are then integrated into the optimization through reprojection constraints
$$
u_{t,j} \sim \pi(K, C_t, X_j),
$$
where $\pi$ denotes the pinhole projection with radial–tangential distortion and intrinsic parameters denoted by $K$.
The residual for observing world point $X_j$ at time $t$ hence is
$$
r_{t,j}^{\text{rep}} = u_{t,j} - \pi(K, C_t, X_j),
$$
which enters the factor graph in the same form as classical bundle adjustment. 
Besides serving as constraints in the optimization, these anchors also allow evaluating the initial reprojection error of a given calibration (see \Cref{fig:introduction_overview}b)). \par

In addition to the reprojection residuals above, anchor uncertainties are modeled on the world-point side. 
For the horizontal plane, we approximate
\[
\sigma_x^W=\sigma_y^W \approx \sfrac{1}{N_{\text{SAT}}},
\]
with $N_{\text{SAT}}$ denoting the orthophoto resolution in px/m. 
Vertical accuracy, in contrast, cannot be inferred from DEM resolution and is instead approximated by the DEM’s reported vertical accuracy~$\sigma_{\text{DEM},z}$.
Beyond these resolution limits, aerial-to-satellite matching introduces noise on both sides of the correspondence, i.e., in UAV and satellite image coordinates.
To keep reprojection constraints fixed, we treat UAV pixel locations $u_{t,j}$ as accurate observations and model combined pixel location errors into the measured world point $\hat{X}^W_j$.
The resulting anchor covariance is thus
$$
\Sigma_j^{\text{anc}}=\Sigma^{\text{SAT/DEM}}+\Sigma^{\text{matching}}
\;\;\approx\;\;
\mathrm{diag}(\sigma_{X_j}^2 I_3),
$$
which acts as a prior on $X_j$, ensuring that both sources of uncertainty from aerial-to-satellite matching and anchor generation are considered properly in the estimation.
With the variables, priors, and uncertainties in place, we next formulate the factor graph used for optimization.

\subsection{Factor Graph Formulation}
\label{subsection:factor-graph-formulation}

\begin{algorithm}[t]
\caption{Staged SCAR Factor Graph Optimization}
\label{alg:scar-optimization}
\begin{algorithmic}[1]
\REQUIRE Pose priors $\{\hat P^W_{\text{CAM},t}\}$ including initial extrinsics, anchors $\{\hat{X}_j^W\}$, initial camera calibration $K_{\text{init}}$
\ENSURE Refined intrinsics and optimized states $\{C^\ast_t, X^\ast_j\}$
\STATE Optimize camera rotations $\{R^W_{CAM,t}\}$; fix all others 
\STATE Optimize camera translations $\{p^W_{CAM,t}\}$; fix all others
\STATE Optimize landmarks $\{X_j\}$ in XY-plane only; fix landmarks in elevation, and fix all other variables
\STATE Optimize calibration intrinsics $K$, fix all others
\STATE Joint refinement of all variables
\end{algorithmic}
\end{algorithm}

Building on the previous subsections, we now define the factor graph constructed by SCAR.
The graph comprises camera poses $\{C_t\}$, geodetic anchors $\{X_j\}$, and intrinsic parameters $K$.
Measurements enter the graph as: (i) pose priors and estimates $\{\hat P^W_{\text{CAM},t}\}$ from GNSS/INS and initial extrinsics; (ii) anchor priors and estimates $\{\hat{X}_j^W\}$ encoding the uncertainty of satellite/DEM resolution and matching; (iii) reprojection factors that couple $C_t$, $X_j$, and $K$ through anchor constraints; and (iv) initial camera calibration parameters used and optimized within reprojection.
The resulting structure is illustrated in \Cref{fig:scar_framework_factor_graph}.

The optimization goal is to minimize the weighted sum
$$
\min_{\{C_t\}, \{X_j\}, K} 
\sum_t \| r_t^{\text{cam}} \|^2_{\Sigma_t^{\text{cam}}} +
\sum_j \| r_j^{\text{anc}} \|^2_{\Sigma_j^{\text{anc}}} +
\sum_{t,j} \| r_{t,j}^{\text{rep}} \|^2_{\Sigma^{\text{rep}}},
$$
where the residuals $r_t^{\text{cam}}$, $r_j^{\text{anc}}$, and $r_{t,j}^{\text{rep}}$ have been introduced in the preceding subsections. 
Each term is weighted by its covariance, ensuring consistent integration of heterogeneous information sources.
For $r_{t,j}^{\text{rep}}$, despite modeling errors on the world-point side, we foresee small reprojection errors in the pixel domain that are covered by isotrope covariance $\Sigma^{\text{rep}}$.
\par

Large outliers in image-based anchors from image matching can dominate the solution.
To mitigate this effect, all residuals are wrapped in a robust kernel.

Direct joint optimization of all variables is prone to poor convergence due to the strong coupling between rotations, translations, landmarks, and calibration parameters.
To improve stability, we adopt a staged, block-coordinate schedule: at each stage, a single parameter group is optimized while all others are held fixed, and previously updated groups are re-frozen before proceeding. 
We first refine camera rotations, then camera translations, then landmarks (XY only), then intrinsics, followed by a final joint refinement.
This strategy prevents systematic errors (e.g., extrinsic offsets) from being absorbed into sensitive parameters such as intrinsics. The optimization procedure is summarized in \Cref{alg:scar-optimization}. \par

The factor graph yields refined camera intrinsics together with optimized camera poses. To estimate the refined extrinsic calibration from these results, we subsequently perform a dedicated least-squares adjustment.

\subsection{Calibration Refinement}
\label{subsection:calibration-refinement}

To recover the refined extrinsic transformation, we exploit the relation between optimized camera poses and the original GNSS/INS poses.
Systematic corrections observed in the camera trajectory after factor-graph optimization are interpreted as extrinsic misalignment, while stochastic deviations are attributed to measurement noise.

Formally, the refined extrinsic $T_{opt}^{\text{INS}\to\text{CAM}}$ is obtained via a robust least-squares alignment.
For notational clarity, we define the relative pose discrepancy as
$$
\Delta_t(T) \;=\; \operatorname{Log}\!\Big[ (C_t^\ast)^{-1} \, (\hat{P}^W_{\text{INS},t} \, T) \Big],
$$
where $\operatorname{Log}\!: SE(3) \to \mathbb{R}^6$ denotes the Lie-algebra mapping of the transformation residual into a minimal representation, $C_t^\ast$ are optimized camera poses, and $\hat{P}^W_{\text{INS},t}$ are GNSS/INS poses.
The refined extrinsic matrix is then obtained as
$$
T^{\text{INS}\to\text{CAM}}_{opt}
=\operatorname*{arg\,min}_{T^{\text{INS}\to\text{CAM}}}
\sum_{t} \rho\!\left( \big\| \Delta_t(T^{\text{INS}\to\text{CAM}}) \big\|^2 \right),
$$
where $\rho$ is a robust kernel for outlier mitigation.

This procedure ensures that only consistent, long-term corrections are absorbed into the extrinsic parameters.
Random GNSS/INS noise averages out without biasing the solution, while anchor noise and image-matching inaccuracies affect individual poses but do not induce coherent offsets.
As a result, systematic effects are correctly attributed to extrinsic miscalibration and resolved in the least-squares adjustment.

\section{EXPERIMENTS AND DISCUSSION}
\label{section:experiments-and-discussion}

\subsection{Experimental Setup}
\label{subsection:experimental-setup}

This section presents the experimental evaluation of SCAR.
We first describe the datasets and setup used in our study.
We then provide quantitative results and complement them with qualitative visualizations. \par

\begin{figure}[b]
    \centering
    \setlength{\fboxrule}{0.2pt}
    \setlength{\fboxsep}{0pt}
    \subfloat[A, Seg.~1]{%
        \fcolorbox{black}{white}{\includegraphics[width=0.24\linewidth]{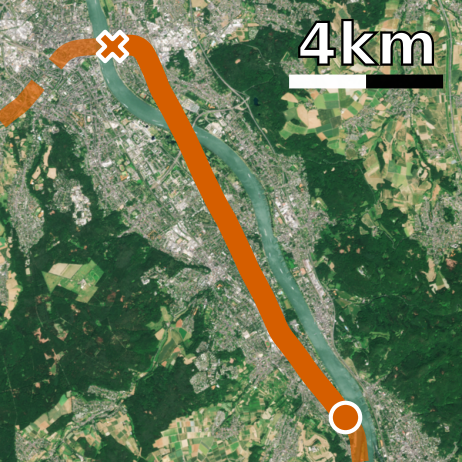}}}
    \hfill
    \subfloat[A, Seg.~2]{%
        \fcolorbox{black}{white}{\includegraphics[width=0.24\linewidth]{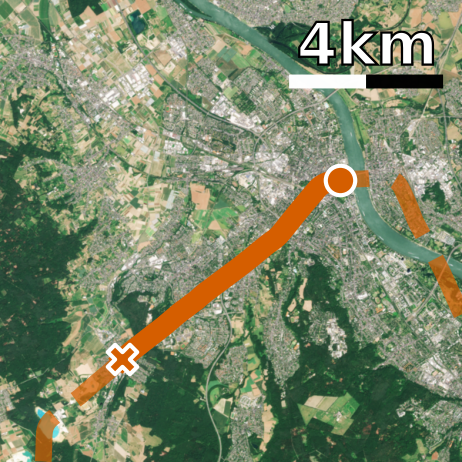}}}
    \hfill
    \subfloat[B, Seg.~1]{%
        \fcolorbox{black}{white}{\includegraphics[width=0.24\linewidth]{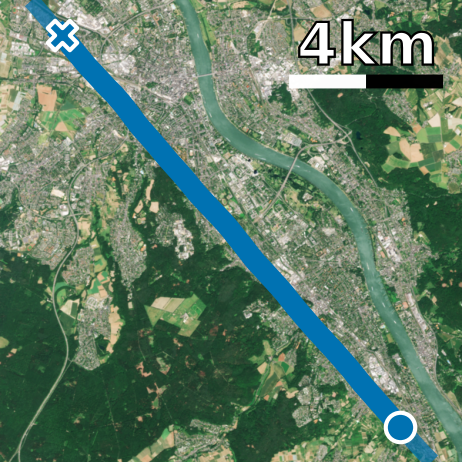}}}
    \hfill
    \subfloat[B, Seg.~2]{%
        \fcolorbox{black}{white}{\includegraphics[width=0.24\linewidth]{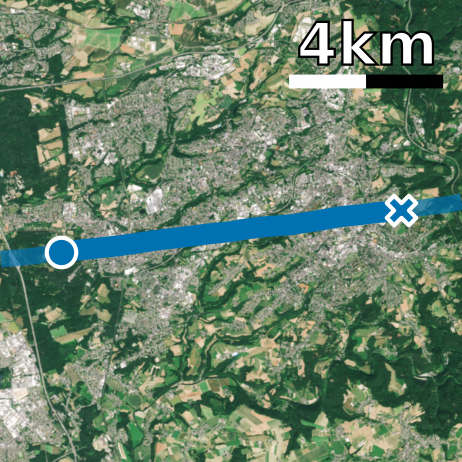}}}
    \caption{
        Selected segments of both routes A (orange) and B (blue).
        For each route, segment 1 is used for calibration, segment 2 for validation.
    }
    \label{fig:calibration-sequences}
\end{figure}

\begin{table}[t]
    \centering
    \caption{Overview of calibration datasets used in the experiments.}
    \label{tab:calibration-datasets}
    \begin{tabular}{lcrrrrr}
    \toprule
    \multirow{2}{*}{ID} & 
    \multirow{2}{*}{Flight Time} & 
    \multirow{2}{*}{\#GCPs} & 
    \multirow{2}{*}{\#Obs.} & 
    \multicolumn{3}{c}{\#Obs./Frame} \\
    \cmidrule(lr){5-7}
    & & & & Mean & Std. & Min.\\
    \midrule
    \multicolumn{7}{c}{\scriptsize Training (1.5k frames each; route/segment A1,\,B1)} \\
    \cmidrule(lr){1-7}
    A1.i   & 2022-02-11 11h & 23{,}798 & 445{,}662 & 296.9 & 58.1 & 81 \\
    A1.ii  & 2022-02-11 15h & 17{,}265 & 368{,}346 & 245.4 & 66.8 & 16 \\
    A1.iii & 2022-06-10 10h & 14{,}946 & 445{,}372 & 296.7 & 66.3 & 56 \\
    B1.iv   & 2022-06-14 10h & 14{,}220 & 463{,}043 & 308.5 & 43.9 & 74 \\
    B1.v  & 2022-10-19 14h & 15{,}965 & 433{,}091 & 288.5 & 47.2 & 120 \\
    B1.vi & 2024-01-29 12h & 15{,}944 & 497{,}359 & 331.4 & 51.9 & 210 \\
    \cmidrule(lr){1-7}
    \multicolumn{7}{c}{\scriptsize Validation (1k frames each; route/segment A2,\,B2)} \\
    \cmidrule(lr){1-7}
    A2.i   & 2022-02-11 11h & 18{,}100 & 334{,}375 & 334.0 & 71.4 & 117 \\
    A2.ii  & 2022-02-11 15h & 14{,}854 & 207{,}742 & 207.5 & 52.7 & 60 \\
    A2.iii & 2022-06-10 10h & 13{,}615 & 285{,}296 & 285.0 & 65.6 & 87 \\
    B2.iv   & 2022-06-14 10h & 10{,}665 & 246{,}853 & 246.6 & 42.0 & 62 \\
    B2.v  & 2022-10-19 14h & 14{,}047 & 221{,}851 & 221.6 & 42.6 & 78 \\
    B2.vi & 2024-01-29 12h & 15{,}234 & 226{,}615 & 226.0 & 36.9 & 87 \\
    \bottomrule
    \end{tabular}
\end{table}

\subsubsection{Datasets}

We evaluate SCAR on a dedicated subset of a forthcoming large-scale multi-season dataset for VIO/VL. This subset will be released as the SCAR benchmark to ensure reproducibility.
The subset consists of six large-scale aerial campaigns recorded between 2022 and 2024 using an ultralight aircraft, representing a long-term deployment of a VIO/VL-system.
Two routes (A and B) were repeatedly flown under varying conditions; the selected segments are illustrated in \Cref{fig:calibration-sequences}.
For each route, we select two spatially consistent subsequences across campaigns: Segment 1 (training, 1.5k frames, $\approx$300\,s at 5\,Hz) for calibration, and Segment 2 (validation, 1k frames, $\approx$200\,s). 
\Cref{tab:calibration-datasets} summarizes the available sequences. 
Each route has been flown three times on different dates, providing six training and six validation sequences in total. \par

The sensor payload comprises a FLIR BFS monochrome camera ($1600\times1100$ px, approx. $60^\circ\!\times45^\circ$ FoV) sampled at 5\,Hz and an SBG Ellipse2-D INS providing GNSS/IMU data. 
All imagery was aligned to publicly available orthophotos and DEMs from North Rhine-Westphalia, Germany.

\subsubsection{Data Preparation}

For SCAR, aerial images are hardware-triggered and synchronized with INS poses.
For each frame, we render a footprint-aligned orthophoto cutout using the INS pose together with the initial camera intrinsics and extrinsics.
SuperPoint~\cite{detone2018superpoint} features are detected in the aerial image and matched to the cutout with LightGlue~\cite{lindenberger2023lightglue}; CLAHE preprocessing and MAGSAC++ homography filtering~\cite{barath2019magsacplusplus} are used to reduce outliers.

To avoid gross elevation errors in the resulting 2D–3D correspondences, we apply a DEM-based slope filter: for each tentative match we query a local DEM neighborhood and reject points whose min–max elevation range exceeds a threshold, i.e., where a small image-plane shift could jump between rooftops and ground.
In addition, we require a minimum pixel spacing between accepted features, which suppresses ambiguous, near-duplicate correspondences.
Across frames, surviving features are tracked with KLT and RANSAC outlier rejection.
For every image, we store anchor IDs with pixel locations, their UTM coordinates, and the synchronized INS pose with uncertainties; these precomputed sequences are the input to SCAR. The factor-graph optimization is then performed using GTSAM~\cite{dellaert2012factor} with a Huber loss.

\subsubsection{VL benchmark preparation}

For an independent, frame-wise VL benchmark, we use the validation segments without tracking.
Each validation frame is matched against a north-aligned $1\times1$\,km satellite cutout centered at the INS position.
Here, we employ RoMa~\cite{edstedt2024roma} rather than LightGlue because, empirically on our data, RoMa is markedly more rotation-tolerant when aircraft heading and map north are misaligned.
Concretely, we run RoMa on four $90^\circ$ rotations of the cutout (2048 matches each), concatenate the matches, perform geometric verification, and estimate the camera pose via PnP.
This benchmark isolates how calibration quality translates into single-image absolute pose recovery; quantitative results are reported in \Cref{subsection:quantitative-results}.

\subsubsection{Baselines}


We compare SCAR against three established calibration pipelines: Kalibr, COLMAP, and the auto-calibration module of VINS-Mono.
These methods differ substantially in their input requirements, and therefore rely on distinct sources of calibration data.
In particular, Kalibr requires recordings of a calibration target (e.g., a checkerboard or AprilGrid) along with sensor motion to excite the system.
We performed such a calibration recording (Oct. 2022) that provides intrinsics $K^{kalibr}_{22/10}$ and extrinsics $T^{kalibr}_{22/10}$.
COLMAP, in contrast, is designed for SfM pipelines and applied to the training segments for auto-calibration of $K^{colmap}$, while extrinsics are fixed from Kalibr.
Finally, the auto-calibration procedure in VINS-Mono operates directly on in-sequence data from the entire trajectory from take-off until the end of the training segment, rather than being initialized within the segment itself, to ensure fair initialization.

\subsubsection{Metrics}
Calibration quality is measured by the reprojection error of georeferenced anchors, defined as
$$
err_{t,j}^{\text{rep}} = \left\| \pi\big(\hat{X}_j^W\big) - u_{t,j} \right\|_2,
$$
and reported as median and median absolute deviation (MAD) for each sequence. 
We additionally evaluate SCAR with our VL benchmark. 
Here, frame-wise poses are estimated via PnP and compared against INS-Groundtruth. 
VL performance is reported as median translation/rotation error and pose accuracy at thresholds 2\,m/2°, 5\,m/5°, and 10\,m/10°. \par

\subsubsection{Ablation and Robustness}

For ablation studies, specific modeling choices are selectively disabled: 
(i) removal of nadir constraints, 
(ii) merging of rotation and translation into a single camera-pose optimization, 
and (iii) skipping the final refinement step. 
To make the impact of these design choices measurable, we intentionally degraded the initial intrinsics and extrinsics to simulate poor initial calibration. 
All ablation variants, next to the core method, were run from these degraded initializations to provide a fair comparison. 

For robustness, we vary the number of training frames included in the optimization process. 
This analysis examines how sensitive SCAR is to reduced data availability and whether stable estimates can be obtained from fewer frames.

Next, we present the quantitative evaluation of SCAR against different baselines. 
We first report calibration quality in terms of reprojection error across training and validation sequences, 
and then assess the impact on VL performance.

\subsection{Quantitative Results}
\label{subsection:quantitative-results}

\begin{table}[tb]
    \centering
    \caption{VL-based evaluation on validation sequences.}
    \label{tab:results-scar-vl}
    \begin{tabular}{l|l|c|c|rrr}
    \toprule
    \multirow{2}{*}{ID} &
    \multirow{2}{*}{Calib.} &
    \multicolumn{2}{c|}{Pose error} &
    \multicolumn{3}{c}{Pose acc.\,[\%]\,@\,m/°} \\
     & & Rot. [°] & Trans. [m] & 2/2 & 5/5 & 10/10 \\
    \midrule
    \multirow{2}{*}{\shortstack{A2.i \\ \Vl}} & Kalibr & $2.48 \pm 0.12$ & $2.19 \pm 0.85$ & 0 & 90 & 99 \\ & \textbf{SCAR} & $0.32 \pm 0.13$ & $1.97 \pm 0.79$ & 51 & 93 & 99 \\
    \midrule
    \multirow{2}{*}{\shortstack{A2.ii \\ \Vl}} & Kalibr & $2.64 \pm 0.22$ & $7.23 \pm 2.35$ & 0 & 23 & 75 \\ & \textbf{SCAR} & $0.63 \pm 0.25$ & $4.75 \pm 2.20$ & 14 & 53 & 85 \\
    \midrule
    \multirow{2}{*}{\shortstack{A2.iii \\ \Vl}} & Kalibr & $2.25 \pm 0.18$ & $5.53 \pm 1.80$ & 1 & 41 & 84 \\ & \textbf{SCAR} & $0.58 \pm 0.23$ & $3.78 \pm 1.69$ & 16 & 65 & 88 \\
    \midrule
    \multirow{2}{*}{\shortstack{B2.iv \\ \Vl}} & Kalibr & $2.96 \pm 0.14$ & $4.01 \pm 1.62$ & 0 & 64 & 91 \\ & \textbf{SCAR} & $1.05 \pm 0.11$ & $3.96 \pm 1.87$ & 18 & 63 & 90 \\
    \midrule
    \multirow{2}{*}{\shortstack{B2.v \\ \Vl}} & Kalibr & $1.46 \pm 0.07$ & $3.47 \pm 1.60$ & 20 & 68 & 91 \\ & \textbf{SCAR} & $0.96 \pm 0.10$ & $3.46 \pm 1.73$ & 25 & 68 & 91 \\
    \midrule
    \multirow{2}{*}{\shortstack{B2.vi \\ \Vl}} & Kalibr & $1.72 \pm 0.21$ & $3.81 \pm 3.11$ & 29 & 79 & 95 \\ & \textbf{SCAR} & $0.45 \pm 0.43$ & $2.91 \pm 3.34$ & 30 & 78 & 94 \\
    \bottomrule
    \end{tabular}

    \vspace{0.75ex}
    {\scriptsize
    Pose error values are median$\pm$MAD [°/m]; \scriptsize\textsc{(Vl)} VL benchmark data.
    }
\end{table}

\begin{table*}[t]
  \centering
  \caption{Comparison of calibration baselines and scar on median reprojection error across calibration sequences.}
  \label{tab:results-scar-calibration}
  \begin{tabular}{
        l|
        S[table-format=2.2] @{\;\(\pm\)\;} S[table-format=2.2]|
        S[table-format=2.2] @{\;\(\pm\)\;} S[table-format=2.2]|
        S[table-format=2.2] @{\;\(\pm\)\;} S[table-format=2.2]|
        S[table-format=2.2] @{\;\(\pm\)\;} S[table-format=2.2]|
        S[table-format=2.2] @{\;\(\pm\)\;} S[table-format=1.2]|
        S[table-format=2.2] @{\;\(\pm\)\;} S[table-format=1.2]
    }
    \toprule
    \multirow{2}{*}{ID} &
    \multicolumn{2}{c|}{Kalibr (ref.)} &
    \multicolumn{2}{c|}{COLMAP} &
    \multicolumn{2}{c|}{VINS-Mono$^{(\text{a})}$} &
    \multicolumn{6}{c}{SCAR [\textbf{ours}]} \\
     & \multicolumn{2}{c|}{2022-10-04} 
     & \multicolumn{2}{c|}{($K\opt$; $T\refa$)} & \multicolumn{2}{c|}{($T\opt$; $K\refa$)}
     & \multicolumn{2}{c|}{($K\opt$; $T\refa$)} & \multicolumn{2}{c|}{($T\opt$; $K\refa$)} & \multicolumn{2}{c}{($K\opt$, $T\opt$)} \\
    \midrule
    A1.i \Tr & 45.90 & 7.21 & 45.91 & 7.23 & 13.14 & 4.66 & 46.22 & 7.19 & 5.61 & 2.48 & \bfseries 5.51 & 2.44 \\
    A2.i \Val & 47.16 & 7.50 & 47.14 & 7.52 & 12.76 & 4.39 & 47.44 & 7.52 & 7.19 & 2.83 & \bfseries 7.05 & 2.81 \\
    \addlinespace[2pt] 
    A1.ii \Tr & 47.45 & 8.54 & 47.90 & 8.58 & 18.52 & 6.39 & 47.07 & 8.64 & 9.06 & 3.94 & \bfseries 8.77 & 3.87 \\
    A2.ii \Val & 47.47 & 8.77 & 47.84 & 8.85 & 17.85 & 6.43 & 46.87 & 8.92 & 10.26 & 4.22 & \bfseries 9.36 & 4.05 \\
    \addlinespace[2pt]
    A1.iii \Tr & 46.19 & 7.26 & 46.32 & 7.28 & 30.28 & 6.03 & 46.61 & 7.31 & 7.43 & 3.24 & \bfseries 7.31 & 3.20 \\
    A2.iii \Val & 46.29 & 6.40 & 46.40 & 6.42 & 29.33 & 4.79 & 46.59 & 6.44 & 7.62 & 3.18 & \bfseries 7.46 & 3.15 \\
    \addlinespace[2pt]
    B1.iv \Tr & 44.67 & 6.06 & 45.57 & 6.07 & 11.60 & 4.03 & 45.09 & 6.05 & 5.16 & 2.23 & \bfseries 5.05 & 2.20 \\
    B2.iv \Val & 51.47 & 8.94 & 52.37 & 8.98 & 16.96 & 5.84 & 51.90 & 8.96 & 10.09 & 3.81 & \bfseries 10.03 & 3.78 \\
    \addlinespace[2pt]
    B1.v \Tr & 6.73 & 2.58 & 6.64 & 2.55 & 6.64 & 2.71 & 6.68 & 2.58 & 4.86 & 2.41 & \bfseries 4.72 & 2.35 \\
    B2.v \Val & 12.69 & 4.51 & 12.64 & 4.48 & 12.29 & 4.32 & 12.71 & 4.54 & 9.00 & 3.40 & \bfseries 8.99 & 3.42 \\
    \addlinespace[2pt]
    B1.vi \Tr & 13.00 & 6.45 & 13.53 & 6.89 & 18.48 & 8.35 & 13.07 & 6.52 & 4.90 & 4.60 & \bfseries 4.79 & 4.59 \\
    B2.vi \Val & 15.36 & 7.43 & 15.16 & 7.69 & 19.05 & 9.21 & 14.91 & 7.47 & 5.24 & 4.53 & \bfseries 5.06 & 4.45 \\
    \bottomrule
  \end{tabular}

  \vspace{0.75ex}
  {\scriptsize
  $\star$ optimized by the resp. method; $\dagger$ fixed from Kalibr;\,
  \scriptsize{(a)} auto-calib. mod.;\,
  Values are median$\pm$MAD reprojection error [px];\,
  \scriptsize\textsc{(Tr)} training, \scriptsize\textsc{(Val)} validation.
  }
\end{table*}

\begin{figure}[tb]
    \centering
    \setlength{\fboxrule}{0.2pt}
    \setlength{\fboxsep}{0pt}
    \subfloat[A2.ii (Kalibr 22-10-04)]{%
        \fcolorbox{black}{white}{\includegraphics[width=0.48\linewidth]{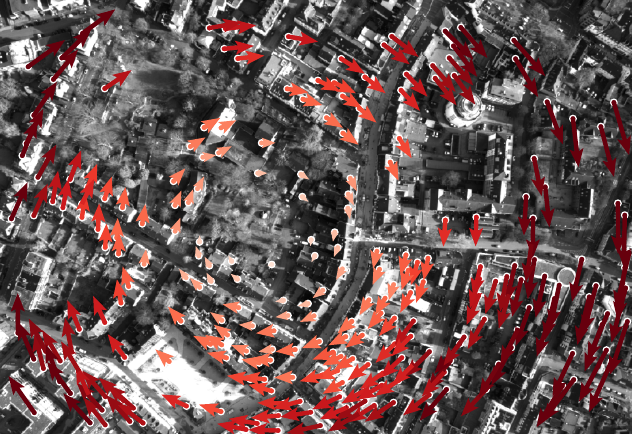}}}
    \hfill
    \subfloat[A2.ii (SCAR)]{%
        \fcolorbox{black}{white}{\includegraphics[width=0.48\linewidth]{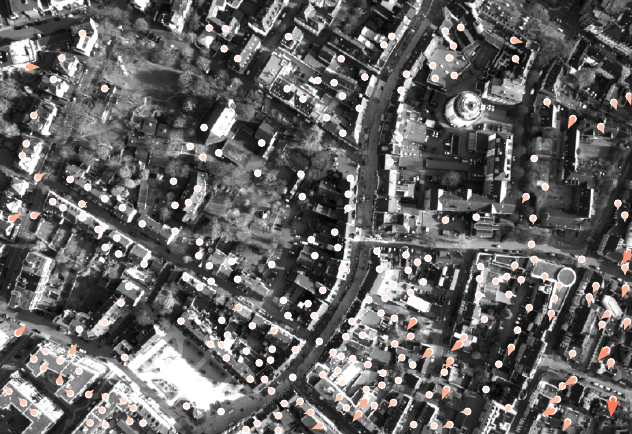}}}
    \\
    \subfloat[B2.v (Kalibr 22-10-04)]{%
        \fcolorbox{black}{white}{\includegraphics[width=0.48\linewidth]{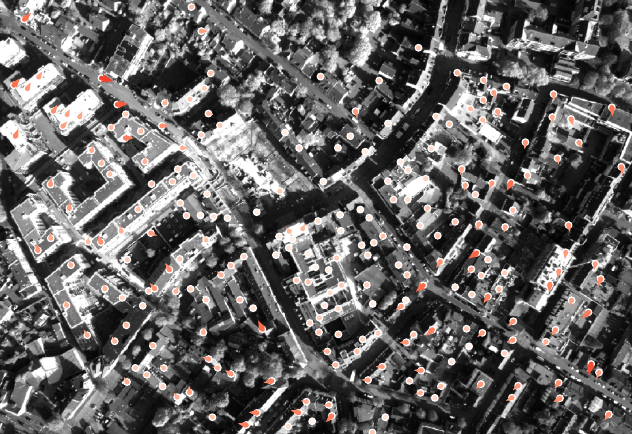}}}
    \hfill
    \subfloat[B2.v (SCAR)]{%
        \fcolorbox{black}{white}{\includegraphics[width=0.48\linewidth]{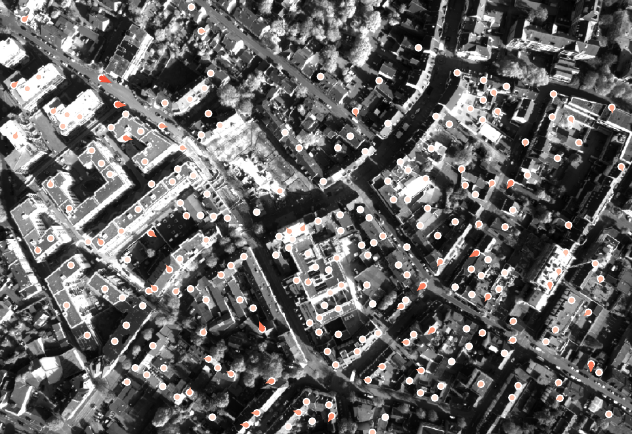}}}
    \caption{Reprojection examples showing reduced errors with SCAR~(b,d) compared to Kalibr~(a,c), with high (a) and low (c) initial error. The colormap in (a,b) ranges from 0 to 100\,px and from 0 to 50\,px in (c,d).}
    \label{fig:qualitative-examples}
\end{figure}

Across all sequences, SCAR consistently reduces the median reprojection error compared to the three baselines: offline calibration with Kalibr, camera-only auto-calibration with COLMAP, and extrinsic-only, IMU-based refinement with VINS-Mono (see \Cref{tab:results-scar-calibration}).
These gains in reprojection accuracy directly transfer to the task-driven evaluation: SCAR achieves substantially lower rotation errors in visual localization and notably higher pose accuracy at strict thresholds 2\,m/2°, and 5\,m/5° (see \Cref{tab:results-scar-vl}). \par

Over all training and validation sequences, SCAR substantially reduces the reprojection error compared to the Kalibr reference.
While COLMAP remains essentially at Kalibr level and VINS-Mono yields selective improvements, SCAR already achieves large gains with $T\opt;K\refa$, and consistently reaches the best performance when using both optimized variables $K\opt$ and $T\opt$ (see \Cref{tab:results-scar-calibration}). 
This confirms that the main source of error in this dataset lies in the extrinsics, yet SCAR with optimized intrinsics still provides an additional benefit in all cases compared to when only optimized extrinsics are used. 
The largest improvements occur in campaigns temporally distant from the Kalibr reference (e.g., for A1.i/A2.i the error is reduced from ca. $45$--47\,px to $\sim 5$--7\,px).
In campaigns closer to the reference (e.g., B1.v/B2.v), the gains are moderate but remain significant, outperforming Kalibr even when a recent high-quality calibration is available. 
Relative improvements persist on validation segments as well, although absolute errors increase, which might stem from slight overfitting on the training data.

The VL evaluation confirms the practical benefit of improved calibration. 
SCAR consistently reduces the rotation error across all sequences (typically from $\sim 2^\circ$ to $\sim 0.3$--$1.3^\circ$) and clearly increases pose accuracy at strict thresholds. 
Translation errors improve more moderately or remain comparable, as extrinsic corrections in translation are only a few centimeters and thus have little effect on the reference pose.  
Rotational corrections, in contrast, are larger and therefore show more significant improvements in VL.
Some sequences (e.g., B2.iv and B2.v) yield nearly identical translation results, yet the rotation improvements persist.

Overall, the improvements achieved by SCAR are both temporally robust (across campaigns from 2022 to 2024 under varying conditions) and spatially consistent (across two routes and their respective segments). 
The relative gains remain stable even when transferring from training to validation data, with the strongest effects observed in campaigns with larger initial errors. 
This highlights the effectiveness of jointly optimizing both $K$ and $T$ under explicit ground-anchor geometry with robust factors, which directly translates into improved VL rotation accuracy.

\begin{figure}[tb]
    \centering
    \setlength{\fboxrule}{0.2pt}
    \setlength{\fboxsep}{0pt}
    \fcolorbox{black}{white}{\includegraphics[height=73pt]{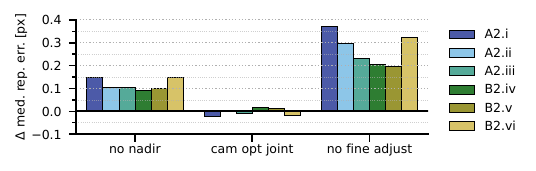}}
    \caption{Ablation study of SCAR. Shown is the change when removing nadir assumptions (“no nadir”), merging camera pose optimization into one step (“cam opt joint”), or skipping the final refinement (“no fine adjust”).}
    \label{fig:ablation-studies}
\end{figure}

\begin{figure}[tb]
    \centering
    \setlength{\fboxrule}{0.2pt}
    \setlength{\fboxsep}{0pt}
    \fcolorbox{black}{white}{\includegraphics[height=73pt]{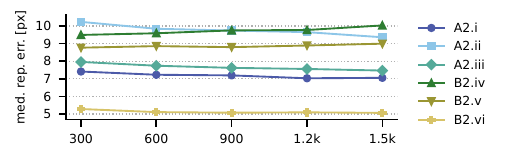}}
    \caption{Robustness evaluation on training data size used for SCAR. Plotted are median reprojection errors on validation segments when calibration is estimated from different numbers of training frames.}
    \label{fig:robustness-studies}
\end{figure}

Beyond the main comparative study, we conducted ablation and robustness analyses to better understand the role of individual components. 
The ablation study (\Cref{fig:ablation-studies}) shows that removing nadir constraints or skipping the final refinement step leads to consistently higher reprojection errors, underlining the importance of these design choices. 
Joint optimization of camera poses in rotation and translation yields errors similar to the core method on average, yet often produces physically implausible extrinsics, confirming the benefit of the staged optimization.
In addition, the robustness analysis (\Cref{fig:robustness-studies}) demonstrates that SCAR remains largely stable when reducing the number of training frames. 
These findings indicate that SCAR maintains stable performance when reducing the number of training frames, with the main sensitivity occurring in cases of poor initial calibration.
We next complement our quantitative findings with qualitative examples, illustrating typical improvements of our method.

\subsection{Qualitative Results}
\label{subsection:qualitative-results}

To qualitatively assess the calibration, \Cref{fig:qualitative-examples} visualizes reprojection errors of automatically generated GCPs. 
This representation makes the impact of calibration improvements directly visible in the image domain. 
We selected two extreme cases: sequence A2.ii with large initial errors under Kalibr, and sequence B2.v with comparatively small errors. 
In both cases, SCAR reduces the reprojection error. 
For A2.ii, Kalibr calibration exhibits a pronounced yaw-related misalignment that is robustly corrected by SCAR. 
In B2.v, where initial errors are already low, a smaller but consistent improvement is still observable. 
In addition to systematic effects, random errors in varying directions remain visible, reflecting noise from the image-based GCP generation.
These outliers are corrected to some degree, but are not part of the calibration evaluation, since only intrinsics and extrinsics are considered as SCAR outputs. 
These qualitative examples show the typical behavior observed across sequences and complement our quantitative evaluation.


\section{CONCLUSIONS}
\label{section:conclusion-and-future-work}


We presented SCAR, a framework for long-term refinement of aerial visual–inertial calibration using automatically generated ground control points from public orthophotos and elevation data. 
Across multiple campaigns and environmental conditions, SCAR proved able to recover from degraded calibrations and consistently improved both reprojection accuracy and downstream visual localization performance. 
These results indicate that long-term operation of aerial systems can be sustained without repeated manual calibration, making large-scale and repeated deployments more practical.  
Current limitations are scene- and location-dependent uncertainties in aerial-to-satellite correspondences, which are presently modeled with global, isotropic covariances rather than per-correspondence uncertainty or correlations along KLT tracks.
This can under-represent local variability.
Accuracy also depends on the quality and recency of orthophotos and DEMs, where vertical DEM errors and outdated imagery may induce parallax, especially around buildings and complex terrain.
Moreover, the method assumes near-nadir views.
Oblique viewpoints or significant roll/pitch reduce geometric robustness and increase ambiguity, potentially degrading calibration.
Finally, mild overfitting can occur when optimizing on specific training segments, leading to slightly higher errors on held-out validation data.
Stronger regularization and cross-region validation can help mitigate this effect.
Addressing these sensitivities and extending the method towards additional sensing modalities form promising directions for future work.  









\bibliographystyle{IEEEtran}
\bibliography{bibliography}

\end{document}